\documentclass[A4, conference]{ieeeconf}  
\IEEEoverridecommandlockouts   
\usepackage{cite}
\usepackage{algorithm} 
\usepackage{algorithmic}  
\usepackage[algo2e]{algorithm2e} 
\usepackage{amsmath,amssymb,amsfonts}
\usepackage{algorithm}
\usepackage{graphicx}
\usepackage{textcomp}
\def\BibTeX{{\rm B\kern-.05em{\sc i\kern-.025em b}\kern-.08em
    T\kern-.1667em\lower.7ex\hbox{E}\kern-.125emX}}
    
    \usepackage{lineno}
\usepackage{lettrine}

\usepackage{url}
\usepackage{moreverb}

\usepackage{graphicx}
\usepackage{subcaption}
\usepackage{float}
\usepackage{amsmath}
\usepackage{array}
\usepackage{multirow}
\graphicspath{ {Figures/} }

\usepackage{scalerel}
\usepackage{tikz}
\usetikzlibrary{svg.path}

\definecolor{orcidlogocol}{HTML}{A6CE39}
\tikzset{
  orcidlogo/.pic={
    \fill[orcidlogocol] svg{M256,128c0,70.7-57.3,128-128,128C57.3,256,0,198.7,0,128C0,57.3,57.3,0,128,0C198.7,0,256,57.3,256,128z};
    \fill[white] svg{M86.3,186.2H70.9V79.1h15.4v48.4V186.2z}
                 svg{M108.9,79.1h41.6c39.6,0,57,28.3,57,53.6c0,27.5-21.5,53.6-56.8,53.6h-41.8V79.1z M124.3,172.4h24.5c34.9,0,42.9-26.5,42.9-39.7c0-21.5-13.7-39.7-43.7-39.7h-23.7V172.4z}
                 svg{M88.7,56.8c0,5.5-4.5,10.1-10.1,10.1c-5.6,0-10.1-4.6-10.1-10.1c0-5.6,4.5-10.1,10.1-10.1C84.2,46.7,88.7,51.3,88.7,56.8z};
  }
}

\newcommand\orcidicon[1]{\href{https://orcid.org/#1}{\mbox{\scalerel*{
\begin{tikzpicture}[yscale=-1,transform shape]
\pic{orcidlogo};
\end{tikzpicture}
}{|}}}}

\usepackage{hyperref} 
\usepackage{caption}

\begin{document}
\title{\LARGE \bf
Swapping Faces, Saving Features: A Dual-Purpose Pipeline for Pedestrian Privacy in ITS
}
    
\author{Roba H. Farouk$^{1,2}$, and Catherine~M.~Elias$^{1,2\orcidicon{0000-0002-1444-9816}\,}$,~\IEEEmembership{Member,~IEEE,}%
\thanks{*This work was not supported by any organization}
\thanks{$^{1}$C-DRiVeS Lab: Cognitive Driving Research in Vehicular Systems, Cairo, Egypt
{\tt\small cdrives.researchlab@gmail.com}}%
\thanks{$^{2}$Computer Science and Engineering Department - Faculty of Media Engineering and Technology - German University in Cairo, Egypt}%
\thanks{{\tt\small roba.ali@student.guc.edu.eg, catherine.elias@ieee.org}}%
}

\markboth{Journal of \LaTeX\ Class Files,~Vol.~14, No.~8, August~2015}%
{author1 \MakeLowercase{\textit{et al.}}:title here}
%



\maketitle
\begin{abstract}
Large-scale and diverse datasets are needed to
train AI models to take real-time decisions for autonomous
vehicles (AVs), an intelligent transportation system (ITS) application. Pedestrian intention and trajectory prediction are critical models used in AVs, requiring datasets involving diverse pedestrian images. Unrestricted access to these datasets imposes serious security risks, like identity theft and pedestrian tracking. The challenge is to apply privacy preservation procedures while maintaining the image attributes needed to train the models. Existing privacy methods may preserve the pedestrian’s privacy, but degrade the image usability, which hinders the models'
effectiveness. This work’s focus is to implement a five-stage pipeline to protect pedestrians’ privacy through face swapping while keeping the essential facial attributes intact. It should be tailored to satisfy the privacy needs of the Egy-DRiVeS dataset. Moreover, Roop and Ghost-v2 face-swapping models are evaluated. Provenly, Roop outperforms Ghost-v2 in various aspects, as will be discussed. Consequently, Roop is the face-swapping model to be used in the pipeline to strike the balance between pedestrian privacy via identity concealment and data usability via facial attribute preservation. 
\end{abstract}



%
\vspace{-10pt}
\section{Introduction and Related Work}\label{sec1}

ITS has contributed to the mobility industry by making systems smarter, safer, and more dynamic. To empower ITS, thriving technologies including AI, global positioning systems (GPS), the internet of things (IoT), and big data must be exploited. Some AI models allow for decision-making abilities and perception mechanisms to be used in AVs, such as pedestrian intention and trajectory prediction and path planning. To train these models, large-scale and diverse datasets are a must. Using AI-based models, the vehicles should be able to predict pedestrians’ anticipated behavior and take appropriate real-time decisions that avoid potential collisions and ensure pedestrians’ safety. Datasets including real pedestrian footage to study their natural behavior in various situations are indispensable. Crucially, facial cues and body movements are the main features needed by these models \cite{intention_prediction_1,intention_prediction_2, elkammar2026vit}. The facial region is in critical need of protection, as it reveals identifiable biometric data about the pedestrian. Unauthorized access to these datasets imposes serious security threats on pedestrians, including identity theft, personal information extraction through data mining, surveillance tracking, and deepfake generation.

 Some countries have established global privacy regulations, such as the General Data Protection Regulation (GDPR) in Europe and the California Consumer Privacy Act (CCPA). Some developing countries are building their own urban datasets while complying with the legal and ethical privacy preservation guidelines. The Egy-DRiVeS dataset is an Egyptian dataset that includes images and videos taken from the public streets \cite{saadawy2024egy}. These urban datasets need specific privacy preservation procedures to handle their unique nature, such as the various pedestrian appearances, such as veiled women. 

 For the AVs applications to operate ethically, pedestrians' privacy should be protected by applying preservation procedures to the datasets. However, the image usability for training tasks after applying these procedures must be considered too. Blurring is commonly used to obfuscate the pedestrian's face, but the image becomes unusable since the facial attributes are no longer clear and recognizable. Thus, privacy preservation techniques that assure pedestrians' privacy while maintaining the data usability for subsequent training tasks are a must.

This work addresses the challenge of preserving pedestrians’ privacy in ITS datasets while ensuring the data usability is maintained. Hence, this work’s aim is to implement a privacy-preserving pipeline that protects pedestrians’ identities and maintains data usability. Focusing on the facial region, face-swapping techniques are used to obscure the pedestrian’s identity while preserving the facial attributes necessary for effective data usability.

Simple image processing techniques such as blurring and pixelation obscure pedestrian identity \cite{traditional1,traditional2}. However, these techniques degrade the image quality, as the facial attributes are unrecognizable and unusable in subsequent model training. Reverse transformation methods \cite{rev-traditional}, such as de-blurring, can retain the original images after processing. Hence, these methods' output neither guarantees privacy nor data usability. PRO-Face framework \cite{pro-face} aims to have pedestrians unidentifiable by humans but identifiable by recognition machines. Obfuscation, as blurring, is initially applied to the facial image; then, the obfuscated image is fused with the original one by a Siamese network. The output is obfuscated, yet still recognizable by traditional face recognizers. While suitable for applications as surveillance videos, where all faces will be anonymized, the system is able to identify the pedestrians for any required task. So, it cannot be used for pedestrian intention prediction since key facial attributes are not visible. 

 Deep-privacy \cite{deep-privacy}, CIAGAN \cite{ciagan}, and Deep-privacy2 \cite{deep-privacy2} models preserve facial privacy through anonymization by using Generative Adversarial Networks (GANs) \cite{gan}. These models generate synthetic faces that conceal the original identity, and they utilize conditional GANs \cite{cgan} to control facial attributes in the new image, such as pose or facial expressions. Although all models achieved high privacy standards, they exhibit low attribute preservation abilities. Thus, they are unsuitable for AVs training datasets that rely on facial cues.

 Face swapping obscures the target's identity by transferring the identity features of another source face to the target while keeping the target's important facial features intact. This approach is promising, as the target face is kept in the frame; the natural distribution and facial structure are unaltered. Thus, face-swapping techniques surpass other methods that generate new anonymized faces in the ability to both preserve facial attributes and protect the pedestrian's privacy. The 3PFS framework \cite{3pfs} presents a pipeline that simultaneously preserves facial attributes used in subsequent tasks and protects pedestrian privacy. The pipeline includes pedestrian and face detectors, a quality enhancer, a source face selection algorithm, and a face swapper. A fine-tuned version of the Ghost face-swapper was used. Later, a new version of the ghost model was developed, Ghost-v2 \cite{ghost-v2}, which is tailored to work on in-wild images. This framework presents the closest approach matching the aim of this work. Although not formally published, Roop \cite{roop} is one of the strongest face swapping models that utilizes a sequence of pretrained models to produce high-fidelity and highly realistic results. The model stages include face detection and analysis, face swapping, and enhancement.

 Based on the reviewed literature and to the best of our knowledge, few methods explicitly address the issue of pedestrians’ privacy in the context of AVs datasets. Most of the existing models work on high-quality facial images, which are uncommon in low-resolution AVs datasets. However, some initiatives address this gap, like Ghost-v2 \cite{ghost-v2} model and the 3PFS pipeline \cite{3pfs}. This highlights the necessity of devoting effort towards filling the existing gap.

 Seemingly, face swapping presents the fine balance required between data usability and privacy. Based on successful multi-stage approaches like the 3PFS pipeline \cite{3pfs} and Roop \cite{roop} model, some pre- and post-processing steps are essential for effective face swapping. Consistently, the main contributions of this work are to:

\begin{enumerate}

    \item Implement a five-stage pipeline that preserves pedestrians' privacy and maintains data usability in the AVs dataset, specially tailored to handle some unique cases found in the urban Egyptian images.

    \item Utilize Roop and Ghost-v2 face swappers for pedestrian de-identification.

    \item Evaluate the face-swapping models in terms of privacy preservation and usability maintenance.

\end{enumerate}

\vspace{-10pt}
\section{Methodology}\label{sec3}
\vspace{-7pt}
\subsection{Pipeline Overview}
\begin{figure*}[]
\centering
\includegraphics[width=0.95\textwidth]{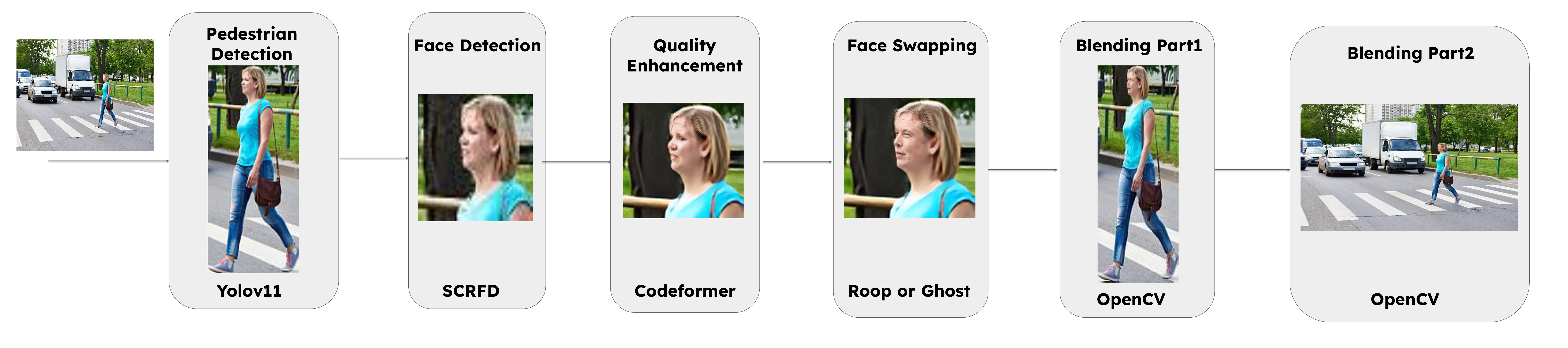}
\caption{The 5 stages of the proposed pipeline.}
\label{fig:full pipeline}
\end{figure*}

 The pipeline is composed of five stages. Firstly, \textit{Pedestrian Detection} detects and classifies pedestrians within a frame. Secondly, \textit{Face Detection} detects the face of the detected pedestrian. Then, \textit{Quality Enhancement} enhances the quality of the detected face by restoration. Then \textit{Face Swapping} is applied, where the facial attributes are transferred from the source face to the target face. Lastly, \textit{Blending} puts the new face back in place of the original one. Facial expressions, head pose, and eye gaze direction are the features preserved during the transfer. As a proof of concept, the pipeline is applied on single frames. The source faces used are fixed and randomly chosen from publicly shareable images. However, the source identity will not be identifiable from the pipeline's output since the source and target faces used will not be revealed, so the process is irreversible. Thus, the source identity is not prone to identification.
Figure~\ref{fig:full pipeline} illustrates the proposed pipeline stages. 

\vspace{-7pt}
\subsection{Stage 1: Pedestrian Detection}
 YOLOV11 pretrained model \cite{yolo11}, one of the latest releases of the YOLO family, is used to detect different objects, encapsulates them in bounding boxes, and annotates them with class names and confidence scores. It achieves higher computational speed, accuracy, and precision. All detected pedestrian objects in the original image are cropped on the boxes coordinates and input to the next stage.
\vspace{-7pt}
\subsection{Stage 2: Face Detection}
 SCRFD pretrained model \cite{scrfd} is used for face detection. Since general object detectors are computationally heavy, SCRFD was optimized to enhance the detection across variant sizes, poses, and even occluded faces, which makes it suitable for AVs datasets. Faces are detected from the cropped pedestrian images and surrounded by bounding boxes, and cropped to be input in the next stage.
\vspace{-7pt}
\subsection{Stage 3: Quality Enhancement}
 Low-resolution images in AVs datasets obstruct the face-swapping procedure, so a quality enhancer is used. General quality enhancers fail to restore facial details required for successful face swapping. Hence, Codeformer \cite{codeformer}, a transformer-based model built for blind face restoration, is used. It uses a codebook to store a set of images for each facial feature to guide the model during restoration, ensuring it chooses the closest features to the face instead of creating new features that may affect facial attribute preservation. 
\vspace{-7pt}
\subsection{Stage 4: Face Swapping}
 The objective of the stage is to ensure the following:
\begin{enumerate}
    \item Identity in the target image is concealed.
    \item Target's facial expressions, eye gaze direction, and head pose are preserved.
    \item Realistic output to be used in further applications.
\end{enumerate}
\subsubsection{Ghost-v2} 
 \textit{Ghost-v2} \cite{ghost-v2} extends face swapping to head swapping, which accounts for more intricate details as face, head shape, and hair, allowing for better attribute preservation. It is tailored to be used on in-wild scenarios that comply with our objective. The architecture is as follows:
\begin{itemize}
    \item \textbf{Aligner}: It performs cross-reenactment by applying the target's pose and facial expressions to the source identity. A set of encoders produces three embeddings for the target's pose, facial expressions, and the source identity at multiple scales. These embeddings are fed into a CGAN to output the source identity with the pose and facial expressions of the target.
    \item \textbf {Blender}: It outputs the final result by using color referencing, inpainting techniques, and binary masking. Color referencing is used for aligning the colors and lighting as in the target image. Masks define the regions to be blended. Finally, inpainting techniques fill and reconstruct missing parts in the final output to match the target image.
\end{itemize}

\subsubsection{Roop}
Although Roop \cite{roop} was developed for entertainment purposes, a closer look onto the internal sequence of pretrained models used explains its suitability for the intended objective. The stages applied are as follows:
\begin{enumerate}
\item \textbf{Face detection}: SCRFD model detects faces from input frames.
\item \textbf{Face analysis}: If there are multiple faces in the input image, facial embeddings are used to choose the closest target face to the source to achieve better realism. Not applicable in our case, as we input single facial images.
    \item \textbf{Face swapping}: \textit{Inswapper\_128.onnx}, a pre-trained face swapping model, is used. It blends the swapped face back into the target image.
    \item \textbf{Face enhancement}: \textit{GFPGAN} \cite{gfpgan} enhancing model is used to improve the image's general quality.
\end{enumerate}
\vspace{-7pt}
\subsection{Stage 5: Blending}
 \textit{SeamlessClone function}, imported from the OpenCV library, is used for blending. The function uses the Poisson image editing technique, which seamlessly blends two images by manipulating the images' gradient vectors. Accordingly, the de-identified image is placed at the target's coordinates, and the pixel values at the boundaries are adjusted to show smooth blending between image gradients, colors, and lighting.
\vspace{-7pt}
\subsection{Evaluation}
 The performance of the entire pipeline is evaluated by assessing the face-swapping models, as this is the core functional part that directly addresses the objective. The other stages of the pipeline are used to prepare the images for face swapping and post-process the swapping output to yield anonymized and usable pedestrian images. Four quantitative metrics are used as follows:
\begin{enumerate}
    \item Landmarks Difference: Facial landmarks represent estimated coordinates of different facial attributes, such as eyes, nose, and mouth corners. 478 landmarks reflect the facial structure, geometry, and pose. The average spatial distance between the original and swapped faces' landmarks is calculated. A low value reflects a well-preserved pose, facial geometry, and structure.
    \item Blendshape Difference: A blendshape predictor uses landmarks to generate 52 facial expression coefficients (e.g., browDownLeft or browDownRight). The sum of the absolute difference between the original and swapped blendshape scores is calculated to represent how much the facial expressions changed. The lower the value, the better the expression preservation.
    \item Facial similarity: Cosine similarity between the original and swapped facial embeddings is calculated, indicating how similar the original and swapped faces are. The lower the value, the more dissimilar are the faces and thus the better the identity preservation.
    \item Gaze vector similarity: Gaze-LLE \cite{gaze} estimates facial gazes; it outputs vectors for the estimated gaze direction. Cosine similarity between the original and swapped vectors is calculated. The higher the value, the better, inferring preservation of target's gaze.
\end{enumerate}

\vspace{-10pt}
\section{Results and Discussion}\label{sec7}
\vspace{-5pt}\subsection{Set 1: Face swapping on high-quality facial images.}
Table \ref{tab:set1} shows the visual output from the face-swapping models on high-quality facial images to asses the quality of the standalone models as a proof of concept.
From table \ref{tab:set1}, it can be noted that:
\begin{enumerate}
\item Both models output realistic images that conceal the target's identity with no artifacts such as ghosting or color gradients mismatch.
\item Both models visually show pose and eye gaze direction preservation —to be quantitatively proved in table \ref{tab:eval}.
\item Roop preserves the head shape, facial structure, and head hair—including eyebrows and beards of the target face. Unlike Ghost-v2, which takes all of these attributes from the source as a result of head swapping.
\end{enumerate}
 The evaluation metrics were calculated using the mean average of a sample of facial close-up images. The following conclusions can be deduced based on table \ref{tab:eval}:
\begin{enumerate}
    \item Roop has lower blendshape difference, so it better preserves facial expressions.
    \item Roop has lower landmark difference, which infers that Roop better preserves the facial structure, geometry, and pose of the target face.
    \item Ghost-v2 has lower cosine similarity, which indicates that the output face is more dissimilar than the original one, better concealing the target identity.
    \item Both have very high gaze cosine similarity values ($\approx 1$) which indicates that the gaze vectors in the original and swapped faces are extremely similar, and thus the attribute is well-preserved.
\end{enumerate}
\vspace{-10pt}\begin{table}[H]
\centering
\caption{Face swapping on close-up facial images.}
\begin{tabular}
{|m{0.21\linewidth}|m{0.21\linewidth}|m{0.21\linewidth}|m{0.21\linewidth}|}
\hline
\textbf{Source} & \textbf{Target} & \textbf{Ghost-v2} & \textbf{Roop} \\ \hline
\includegraphics[width=\linewidth]{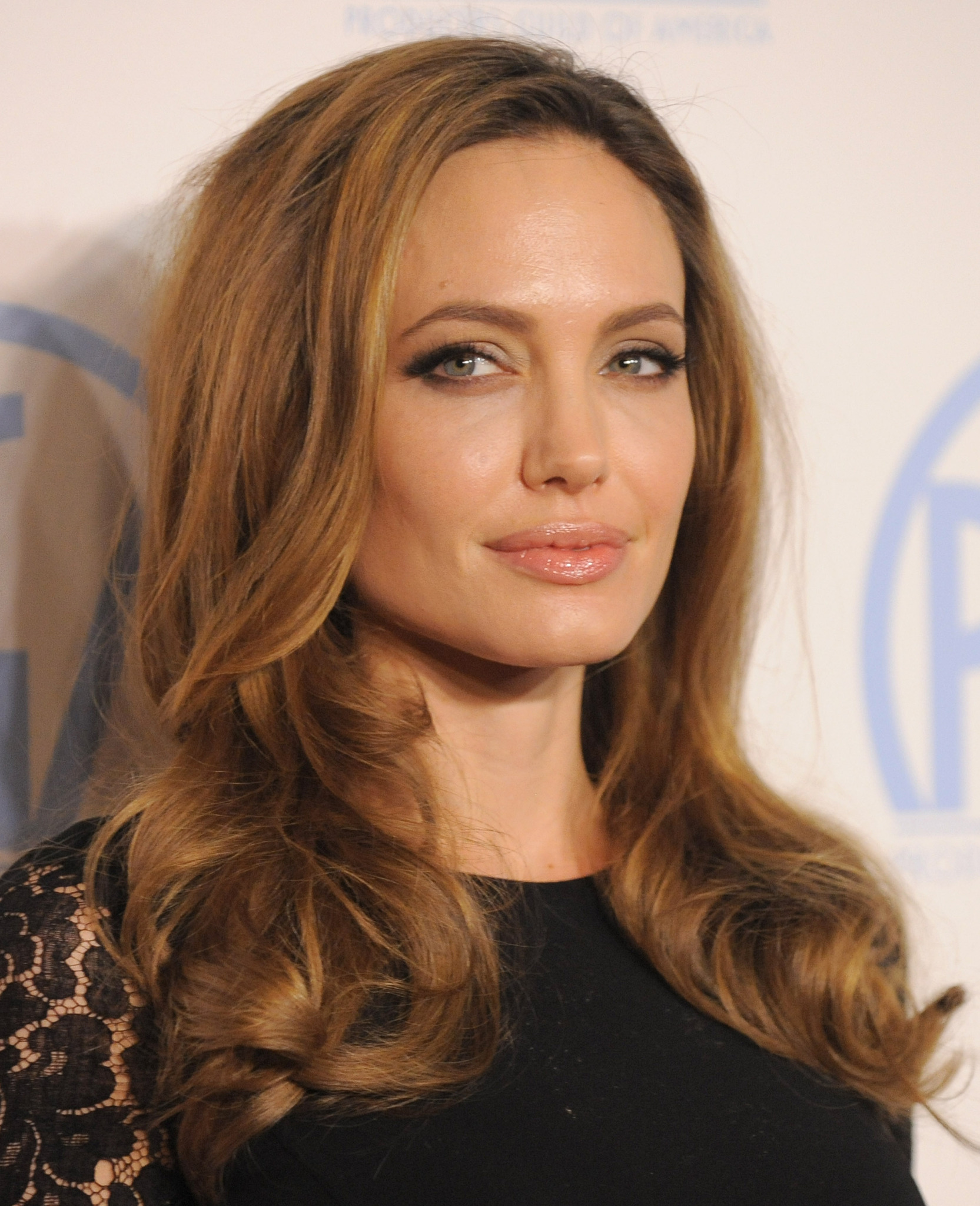} &
\includegraphics[width=\linewidth]{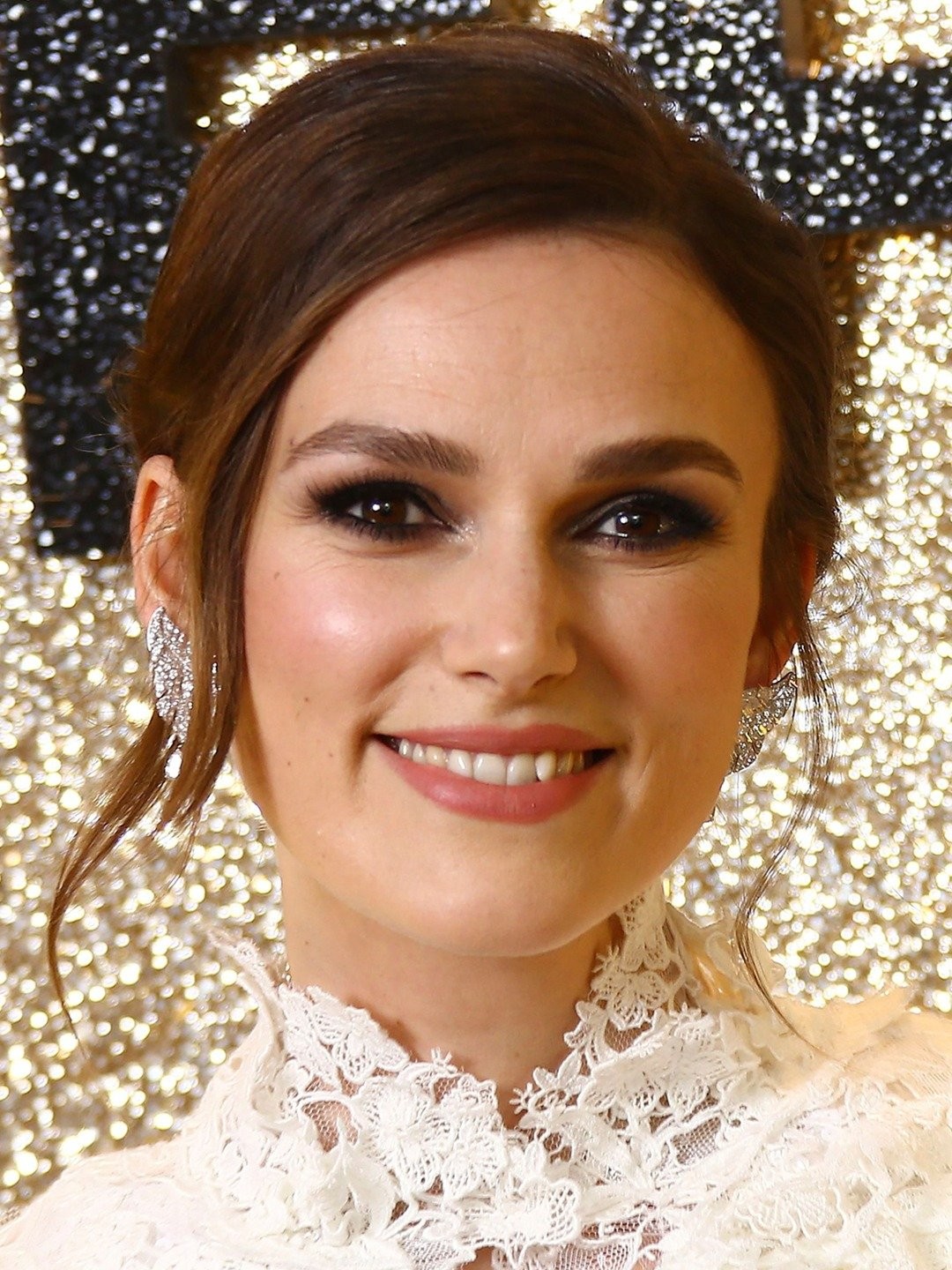} &
\includegraphics[width=\linewidth]{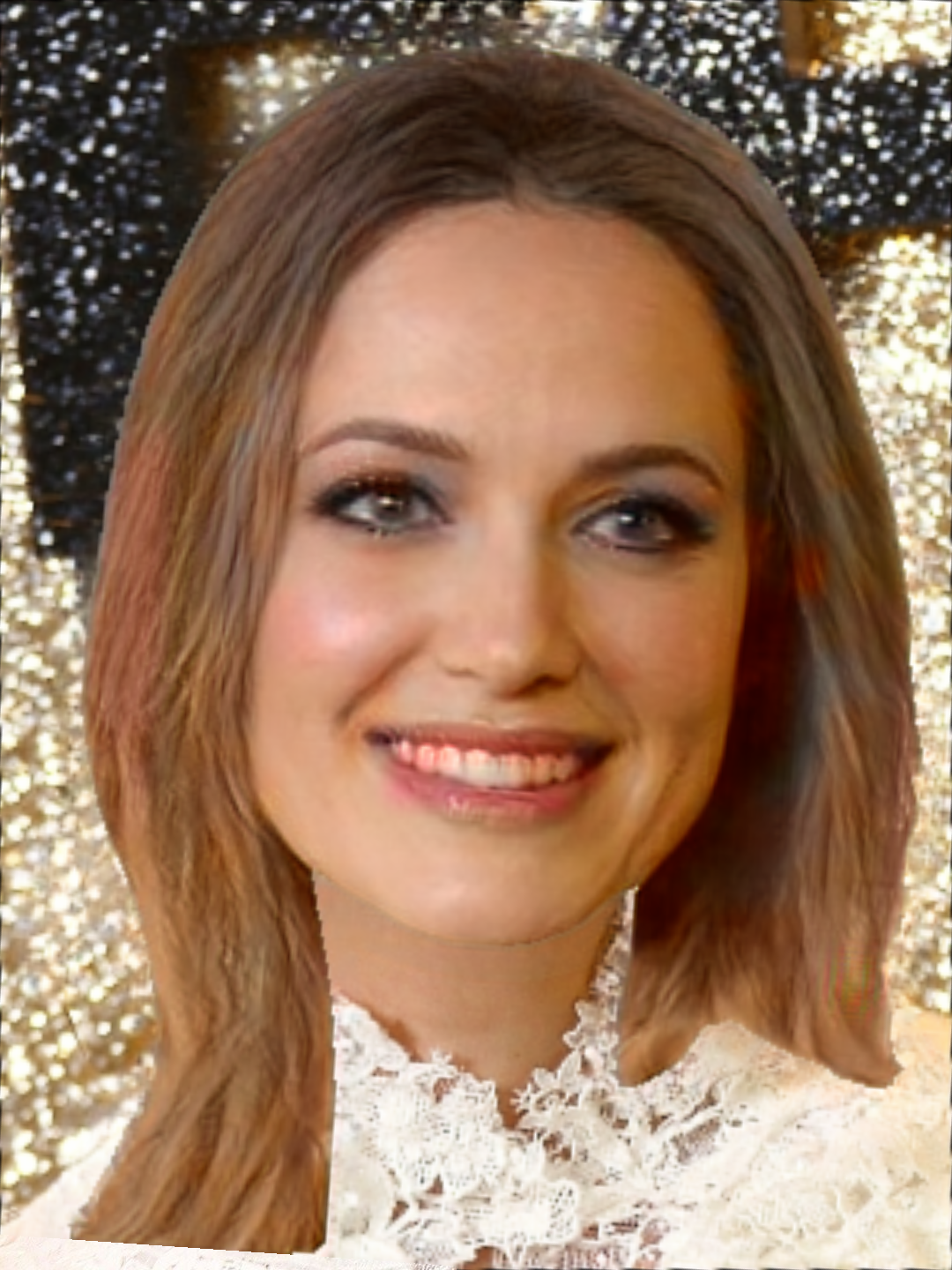} &
\includegraphics[width=\linewidth]{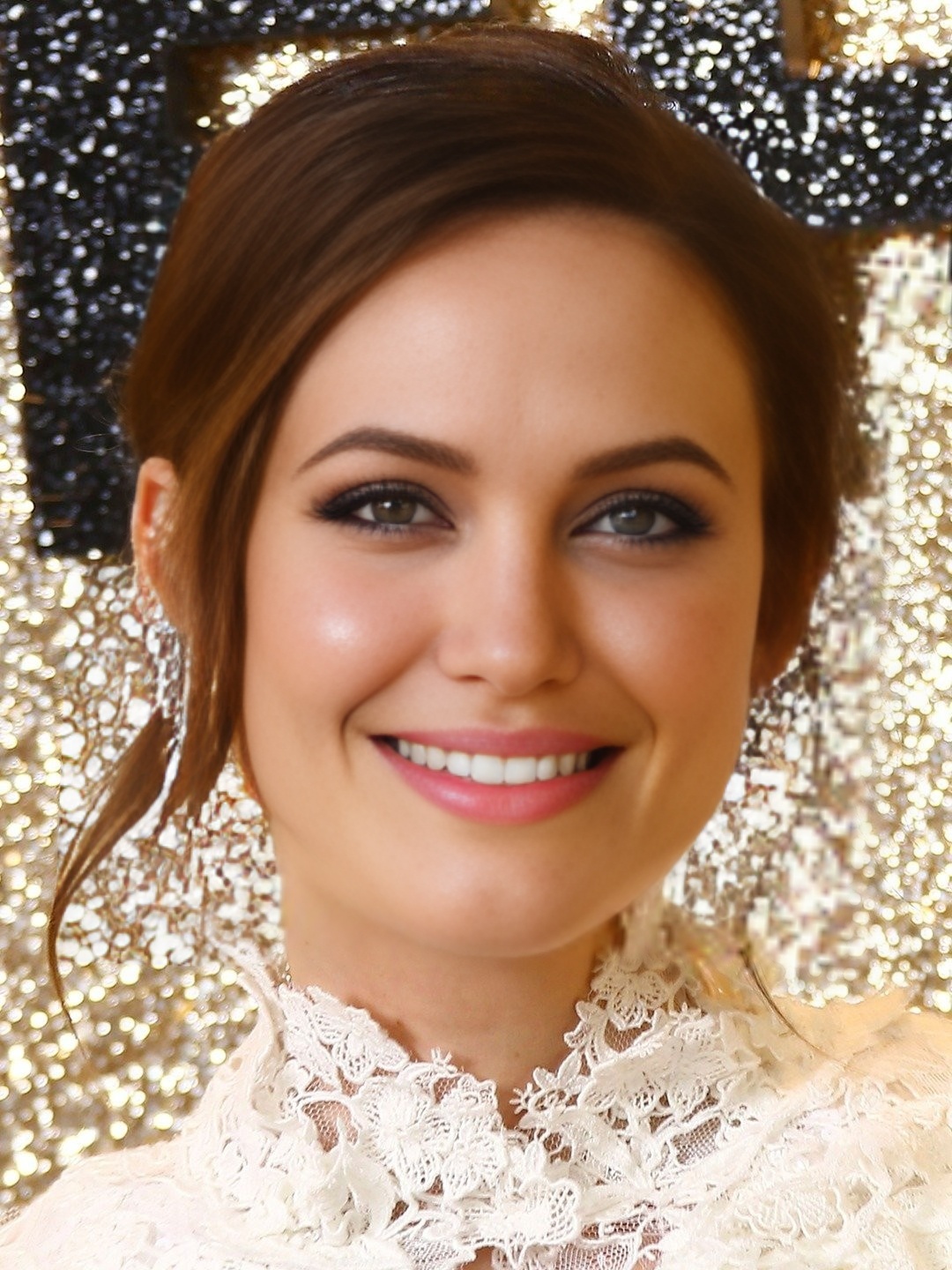} \\ \hline

\includegraphics[width=\linewidth]{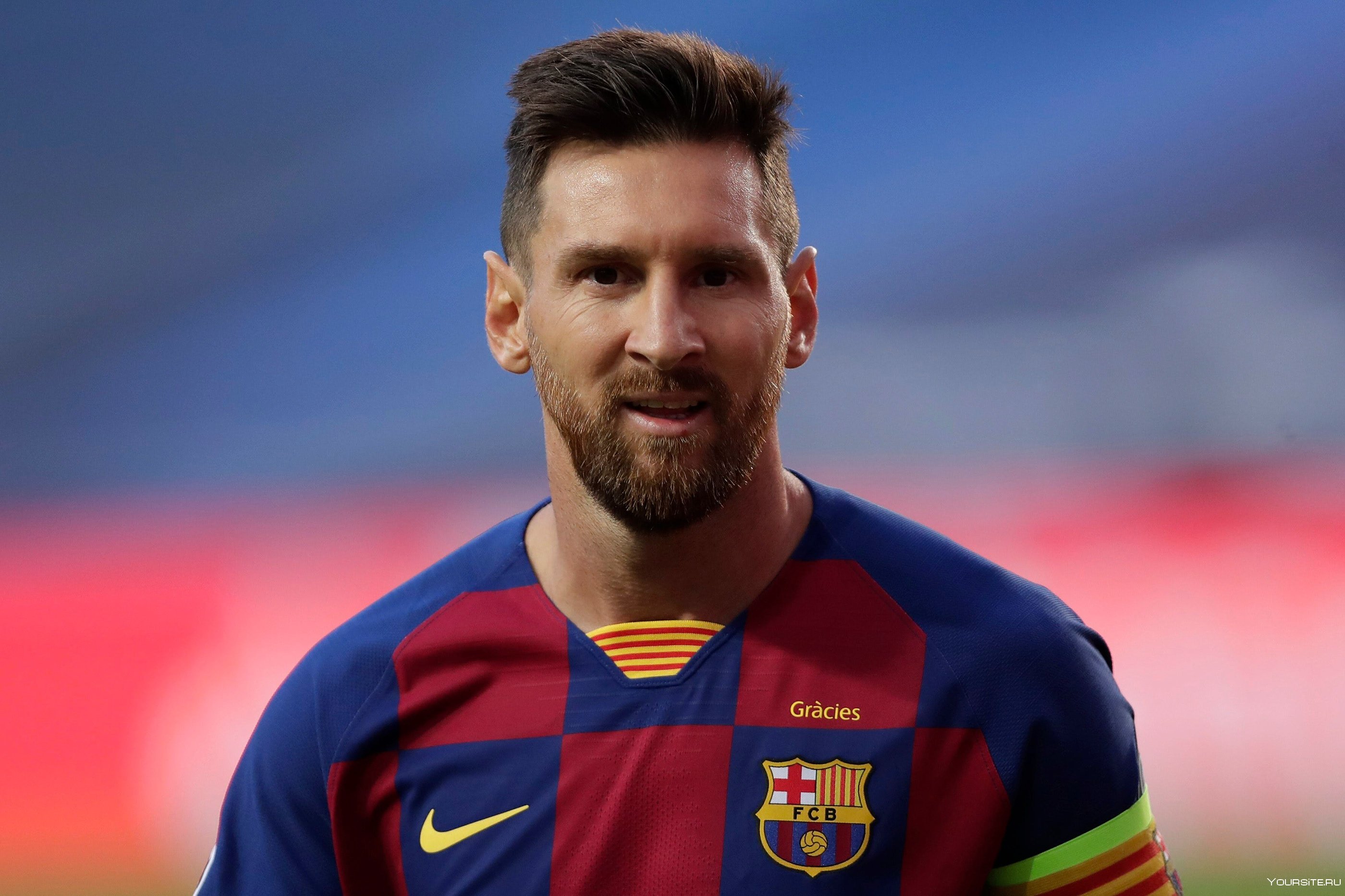} &
\includegraphics[width=\linewidth]{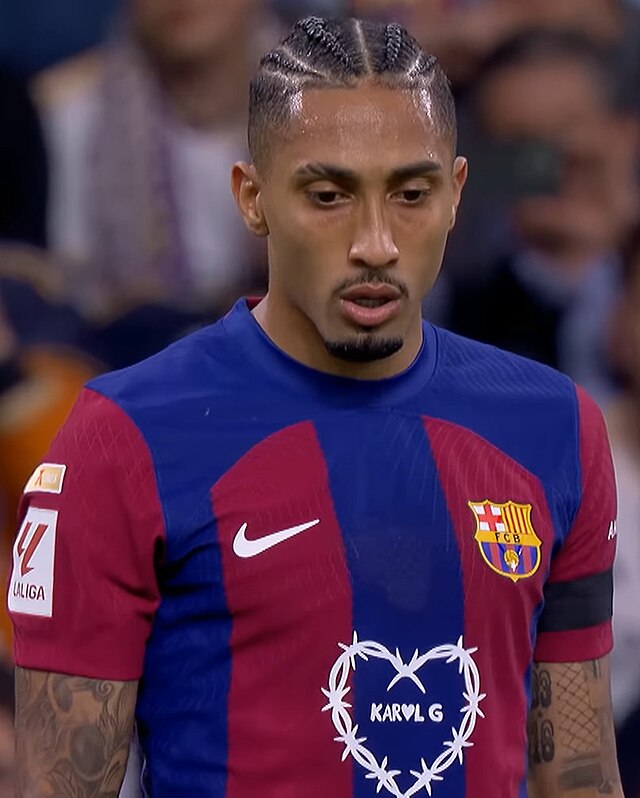} &
\includegraphics[width=\linewidth]{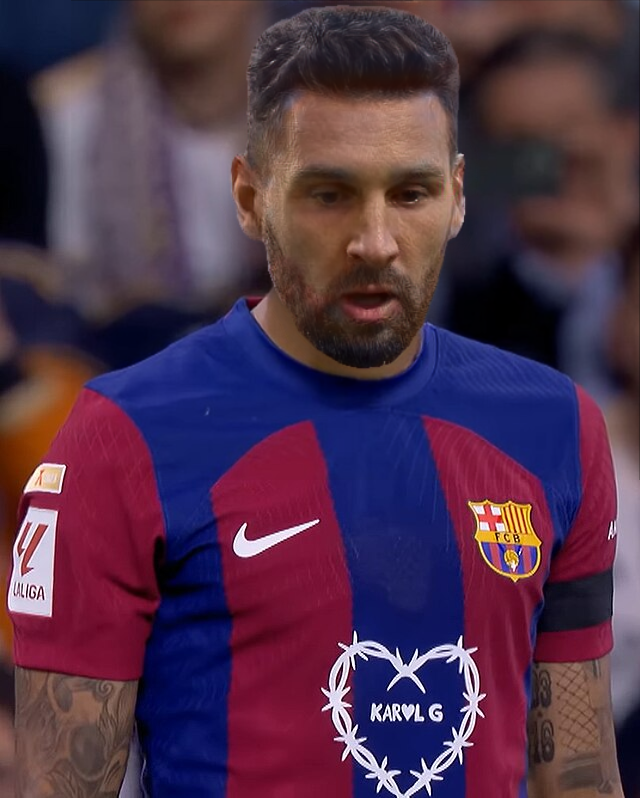} &
\includegraphics[width=\linewidth]{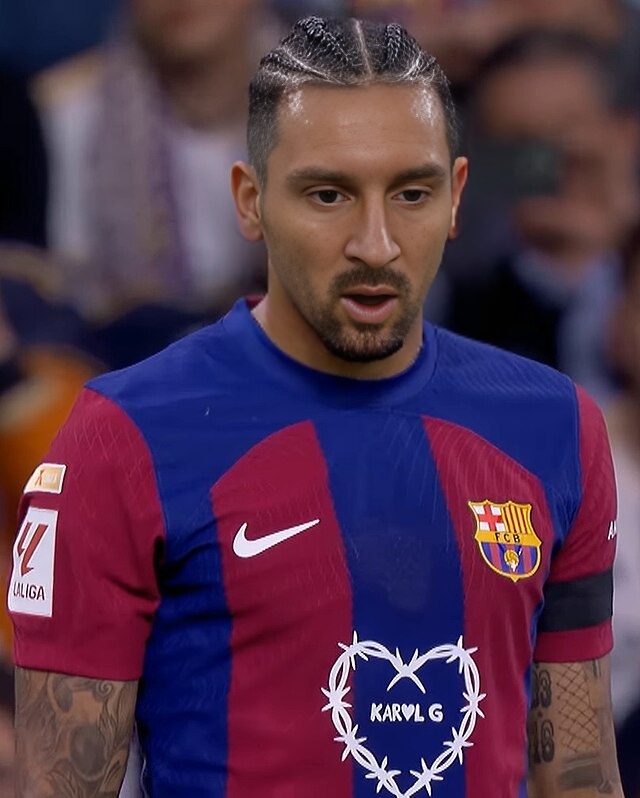} \\ \hline
\end{tabular}
\label{tab:set1}
\end{table}
\vspace{-15pt}\begin{table}[H]
\centering
\caption{Evaluation results of the face swappers.}
\label{tab:eval}
\begin{tabular}{|p{4cm}|p{1.4cm}|p{1.4cm}|}
 \hline
 \multicolumn{3}{|c|}{Evaluation Results} \\
 \hline
 Evaluation Method & Roop & Ghost-v2 \\
 \hline
 Blendshape Difference & \textbf{1.898}
 & 2.0478 \\
 Landmark Difference & \textbf{0.00596}
 & 0.00710 \\
 Cosine Similarity (Identity) & 0.1997
 & \textbf{0.1393 }\\
 Gaze Vectors Cosine Similarity & \textbf{0.9368} & 0.9385\\
 \hline
\end{tabular}
\end{table}

\vspace{-5pt}\subsection{Set 2: Vitality of pedestrian and face detection models.}
\begin{figure}[H]
\centering
\includegraphics[width=0.5\columnwidth]{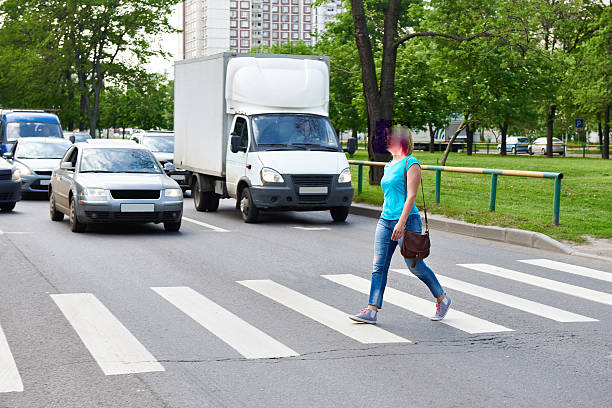}
\caption{Output from Ghost-v2 model on raw street image.}
\label{fig:ghost_fail}
\end{figure}

Figure \ref{fig:ghost_fail} shows that Ghost-v2 was not able to detect the distant face to perform the swap. Although Roop succeeded in most cases due to its internal face detector, it also had a few failures in extreme scenarios, such as very distant pedestrians or extremely distorted faces. As a consequence, pedestrian and face detection modules are used to ensure both models are input clearly detectable faces.

\vspace{-10pt}
\begin{figure}[H]
\centering
\includegraphics[width=0.9\columnwidth]{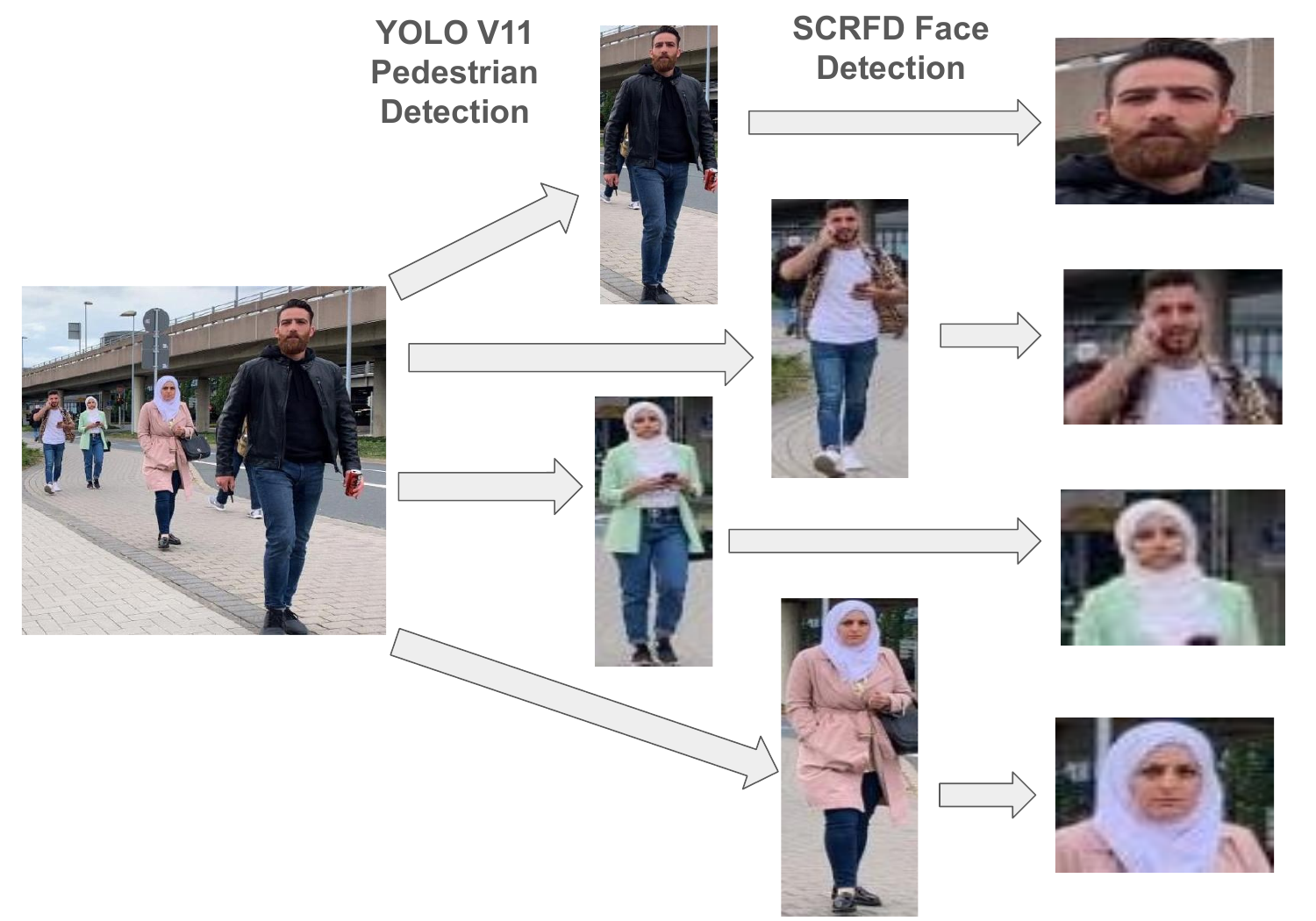}
\caption{Output of the YOLO and SCRFD models.}
\label{fig:col_fig}
\end{figure}

\vspace{-10pt}
Figure \ref{fig:col_fig} shows an input image having multiple pedestrians where all were successfully detected by YOLO model then input to the SCRFD model to have a cropped facial image for every pedestrian at the end of the stage. Real images taken from the Egyptian street were used in all of the following results.

It can be seen from figure \ref{fig:col_fig} distant faces are of low quality and the facial attributes are not visible. Evidently, this obstructed the face swapping process as the models failed to output desired results as shown in table \ref{tab:set2}. Additionally, the following can be deduced from table \ref{tab:set2}:
\begin{enumerate}
    \item Ghost-v2 was not able to handle the occluded face, unlike Roop.
    \item Ghost-v2 is unable to handle veiled woman images; it produced an unrealistic, ghostly face since it applies full head swapping.
    \item Roop proved to be more reliable since it was able to swap faces -unlike Ghost-v2, yet it produced low-quality images that may still be unusable.
\end{enumerate}
\vspace{-10pt}\vspace{-10pt}
\vspace{-10pt}\begin{table}[H]
\caption{Face Swapping results on cropped faces from Egyptian street images.}
\centering
\label{tab:set2}
\renewcommand{\arraystretch}{1.1}
\begin{tabular}{|m{0.20\linewidth}|m{0.20\linewidth}|m{0.21\linewidth}|m{0.21\linewidth}|}
\hline
\textbf{Source} & \textbf{Target} & \textbf{Ghost-v2} & \textbf{Roop} \\ \hline

\includegraphics[width=\linewidth]{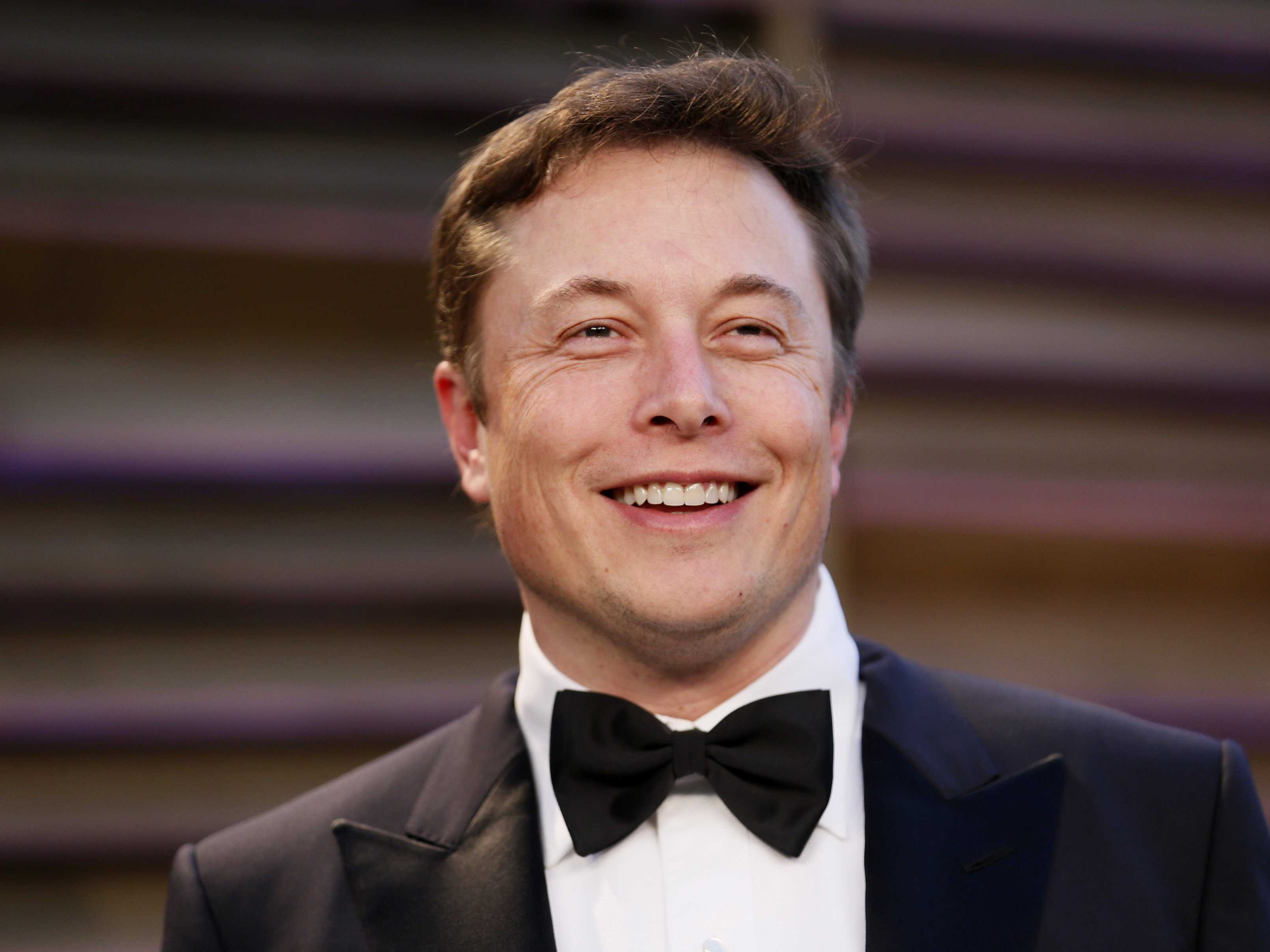} &
\includegraphics[width=\linewidth]{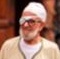} &
\includegraphics[width=\linewidth]{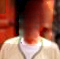} &
\includegraphics[width=\linewidth]{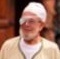} \\ \hline

\includegraphics[width=\linewidth]{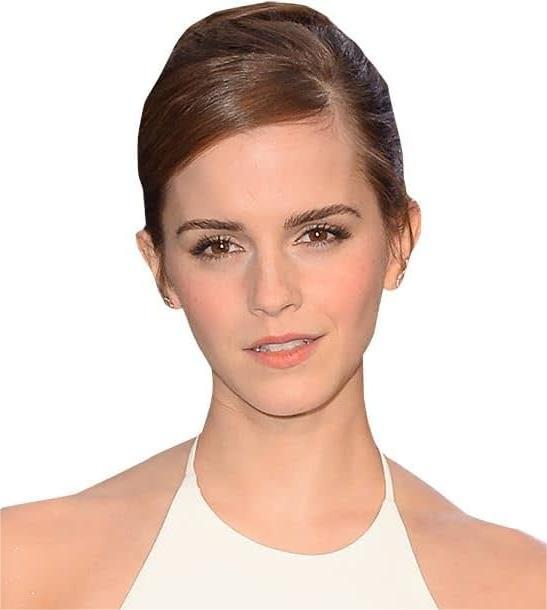} &
\includegraphics[width=\linewidth]{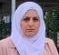} &
\includegraphics[width=\linewidth]{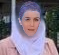} &
\includegraphics[width=\linewidth]{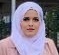} \\ \hline
\end{tabular}
\end{table}

\vspace{-5pt}\subsection{Set 3: Face swapping on quality enhanced images.}
To address the challenge of the low-resolution nature of street images, Codeformer quality enhancing module was applied to the cropped images before being input to the face swapping models. Table \ref{tab:set3} shows the output from the face swapping. The following observations can be noticed:
\begin{enumerate}
    \item Both models' performance enhanced than before using the quality enhancer in terms of ability to output clearly visible swapped of high quality.
    \item Ghost-v2 is still having a problem with images of veiled women in terms of realism, swapping fidelity and even attribute preservation. The output is not visually appealing and the facial expressions are not the same as the target image.
    \item Roop still outperfors Ghost-v2 in terms of output image realism, attribute preservation, and robustness to handle more challenging cases.
\end{enumerate}

\vspace{-10pt}\begin{table}[H]
\caption{Face Swapping results on enhanced cropped faces from street images.}
\centering
\label{tab:set3}
\renewcommand{\arraystretch}{1.2}
\begin{tabular}{|m{0.19\linewidth}|m{0.20\linewidth}|m{0.21\linewidth}|m{0.21\linewidth}|}
\hline
\textbf{Source} & \textbf{Target} & \textbf{Ghost-v2} & \textbf{Roop} \\ \hline

\includegraphics[width=\linewidth]{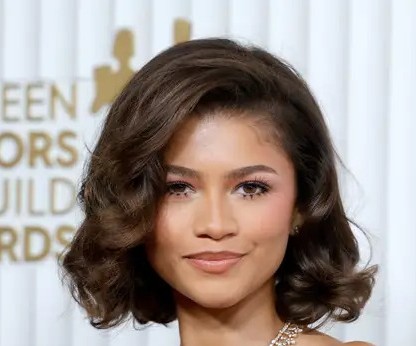} &
\includegraphics[width=\linewidth]{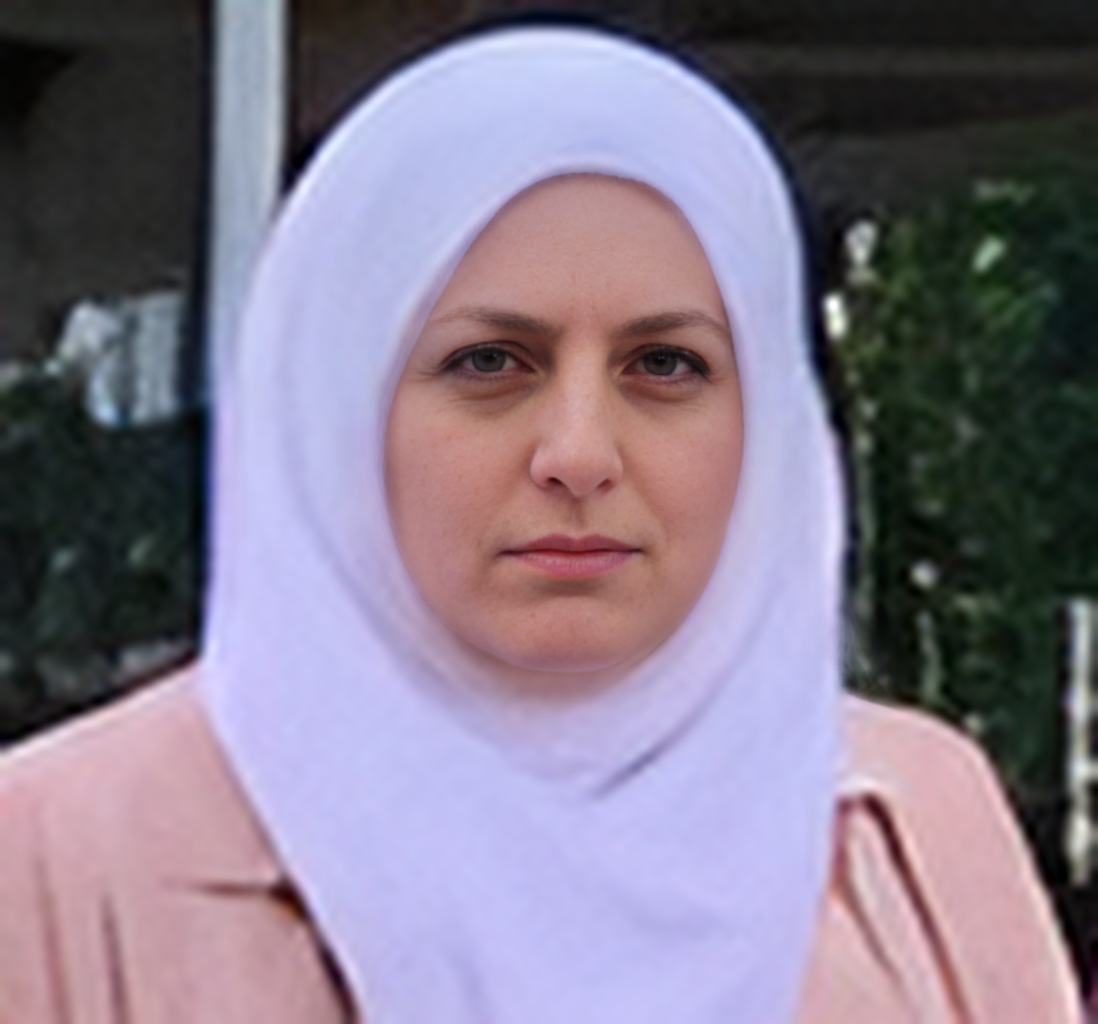} &
\includegraphics[width=\linewidth]{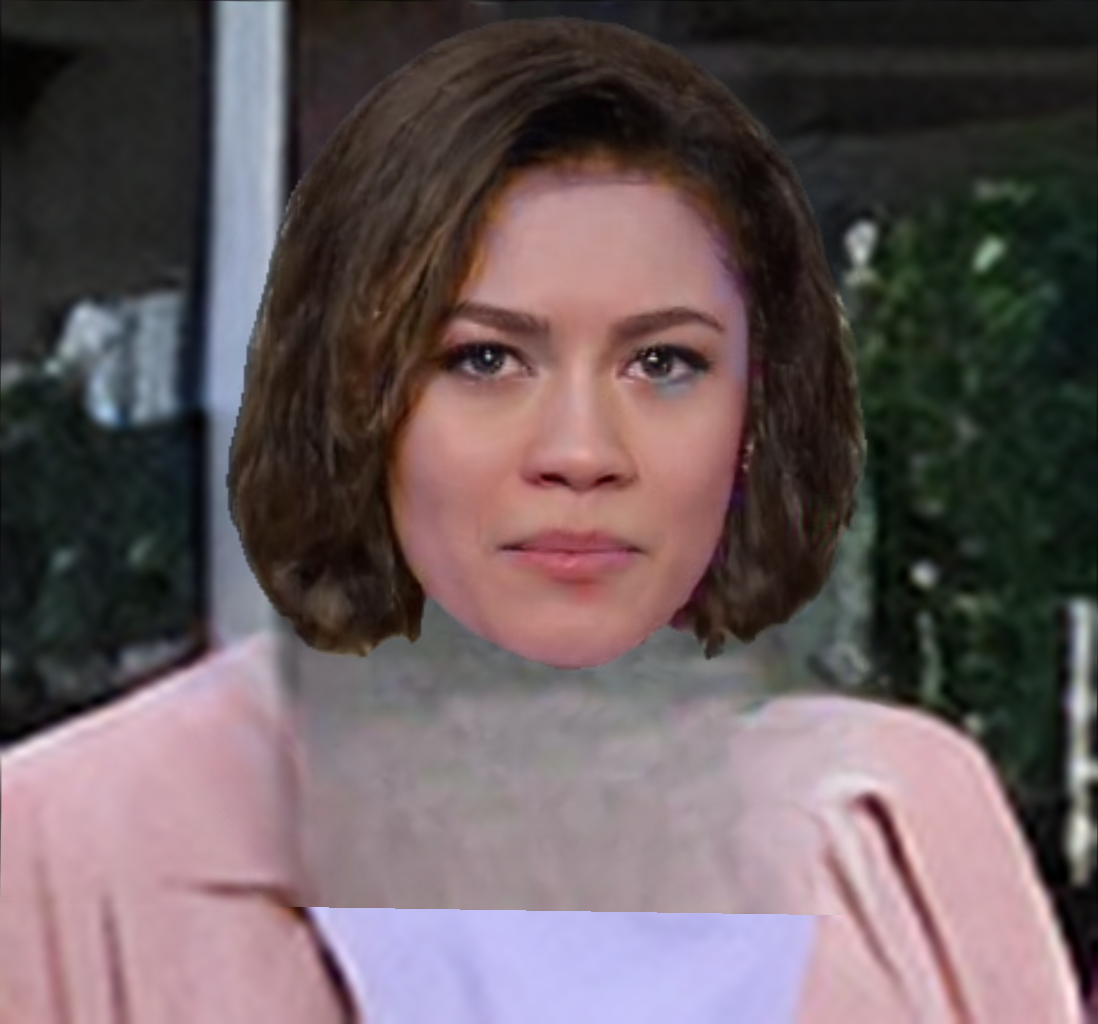} &
\includegraphics[width=\linewidth]{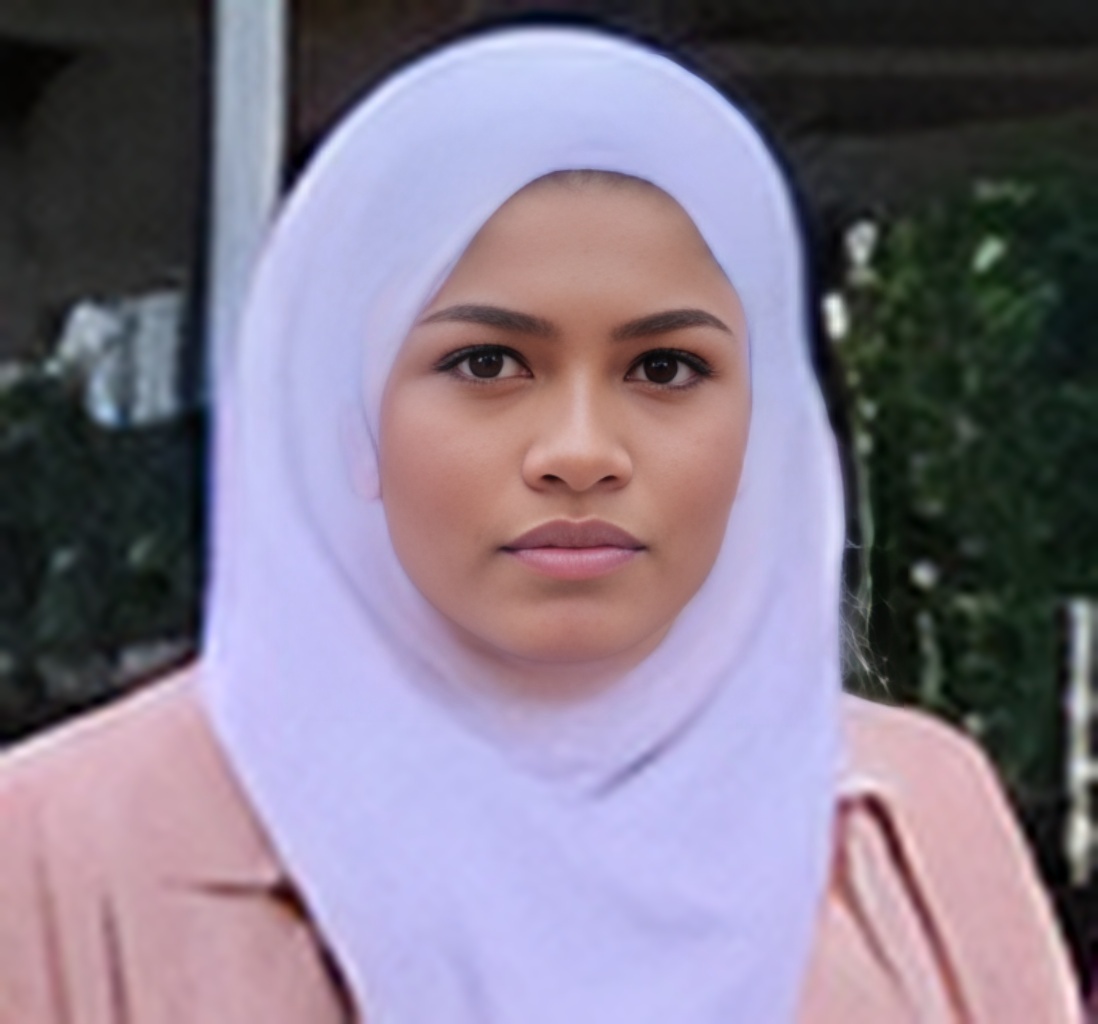} \\ \hline

\includegraphics[width=\linewidth]{Figures/Results/messi.jpg} &
\includegraphics[width=\linewidth]{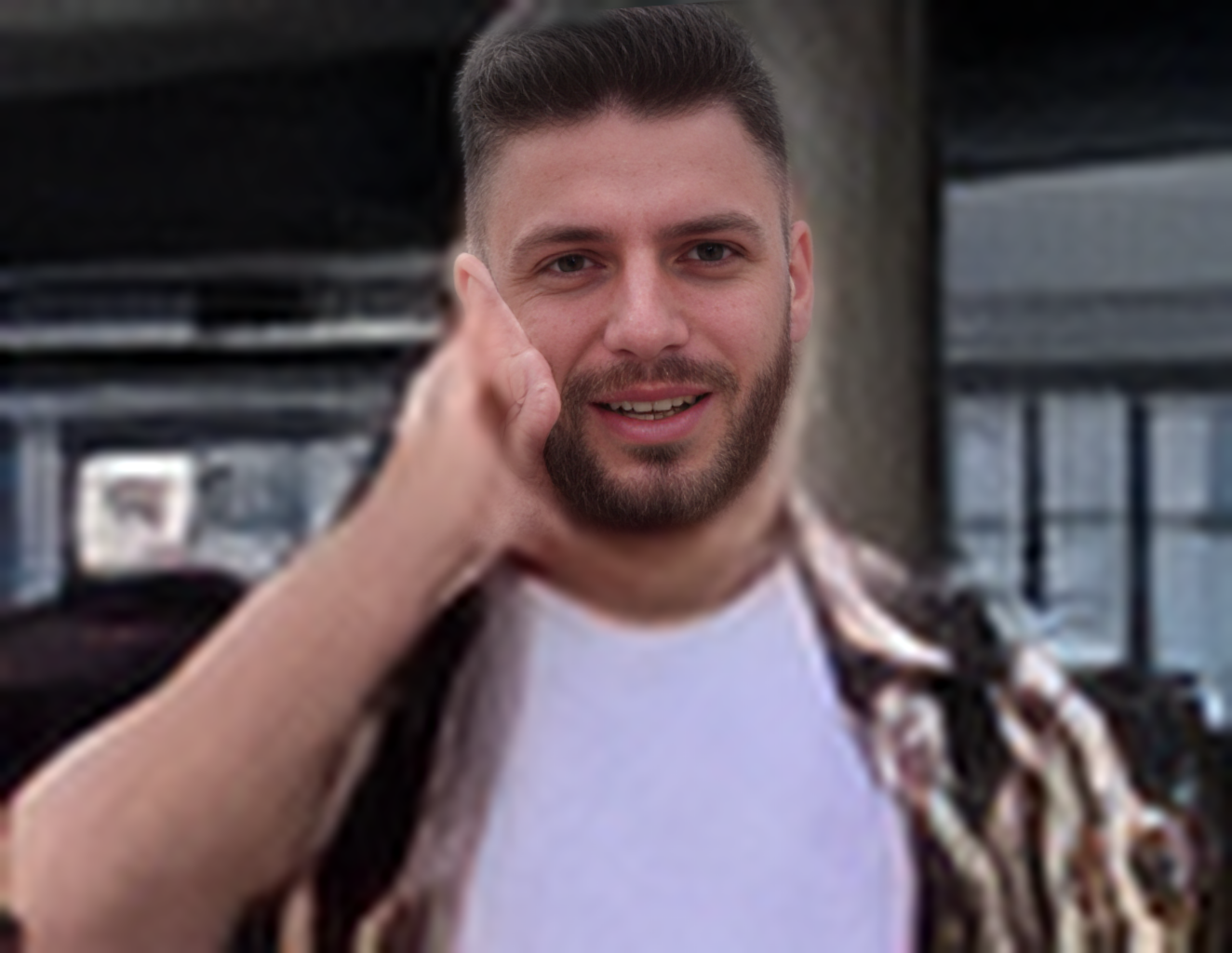} &
\includegraphics[width=\linewidth]{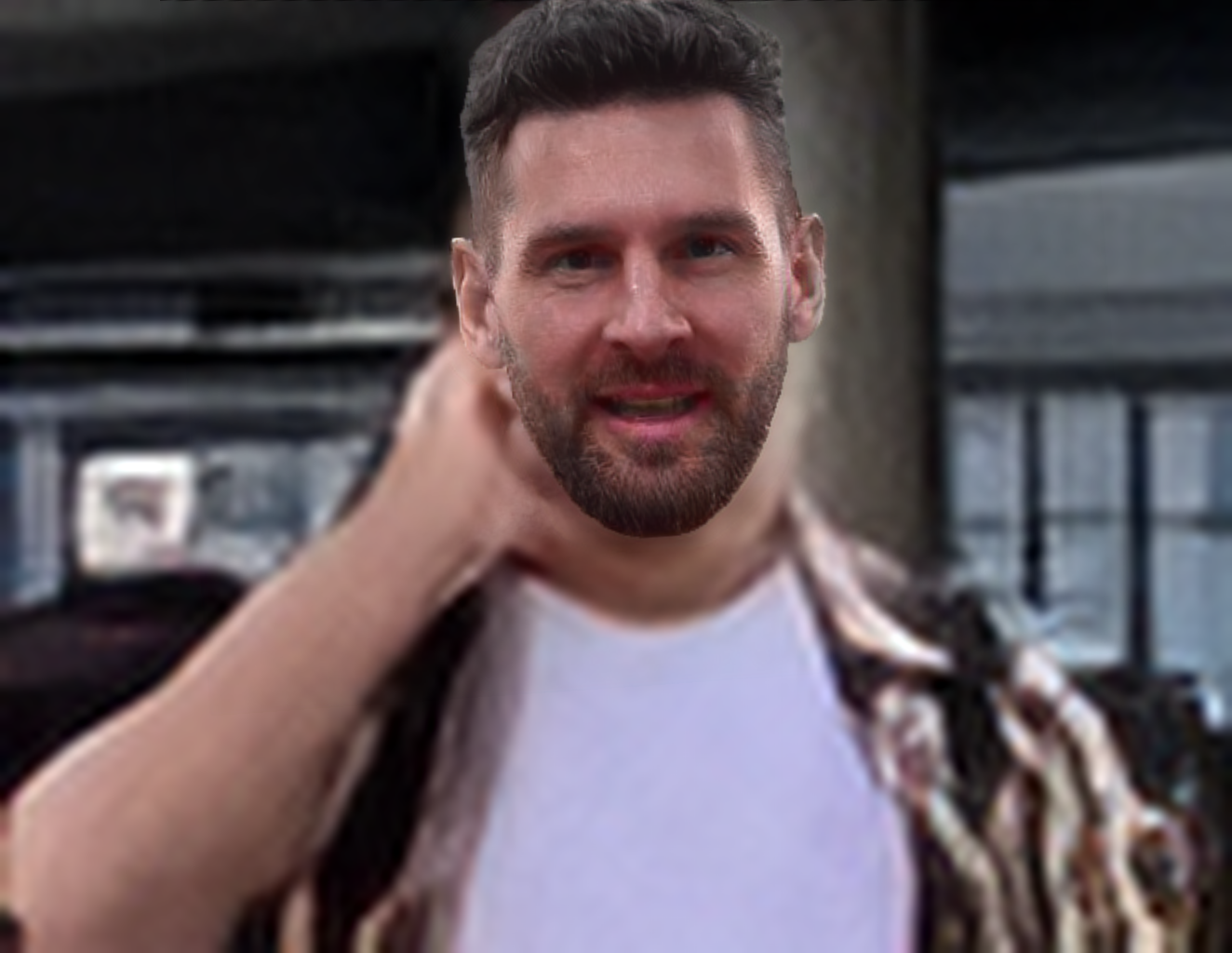} &
\includegraphics[width=\linewidth]{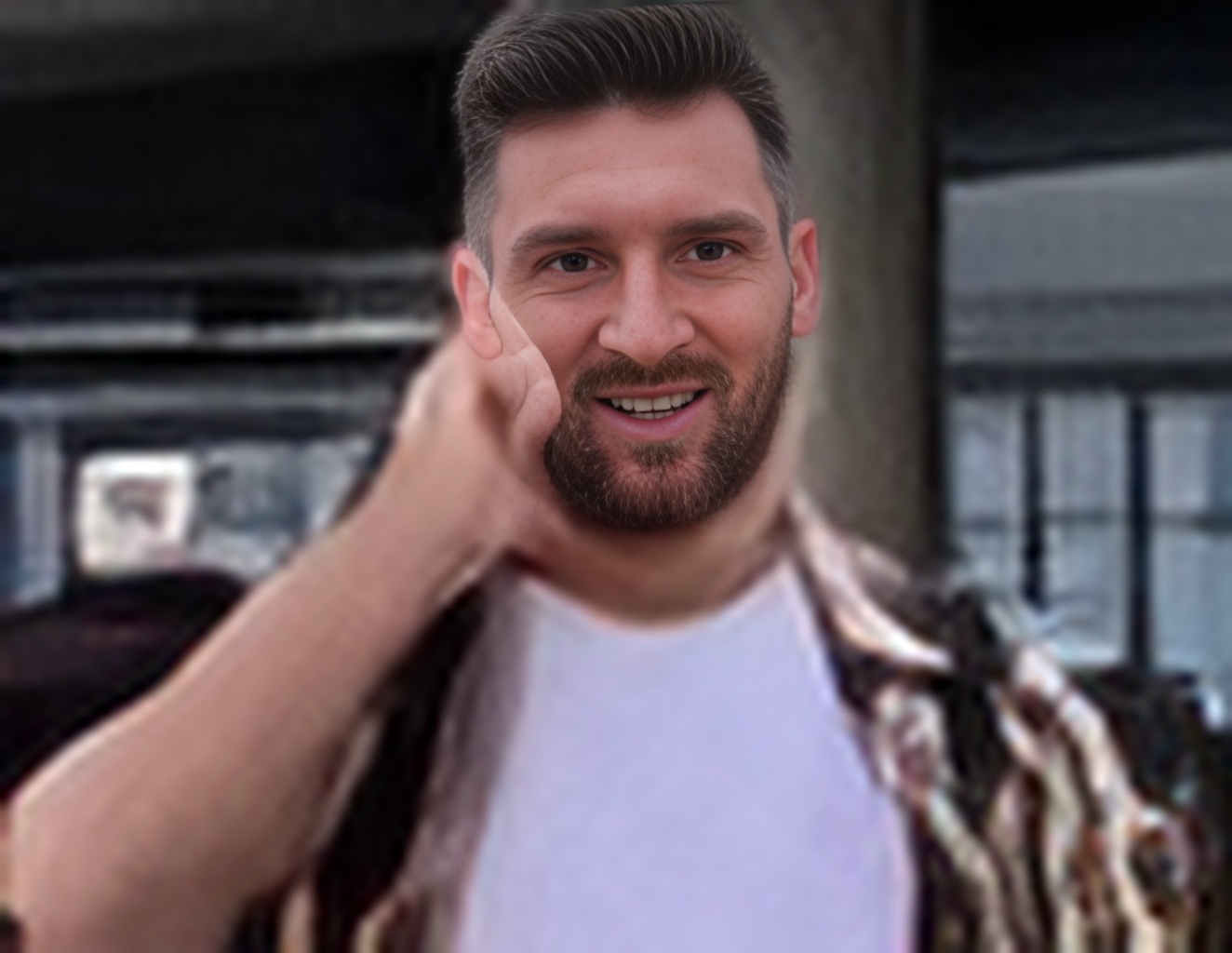} \\ \hline

\includegraphics[width=\linewidth]{Figures/Results/emma.jpg} &
\includegraphics[width=\linewidth]{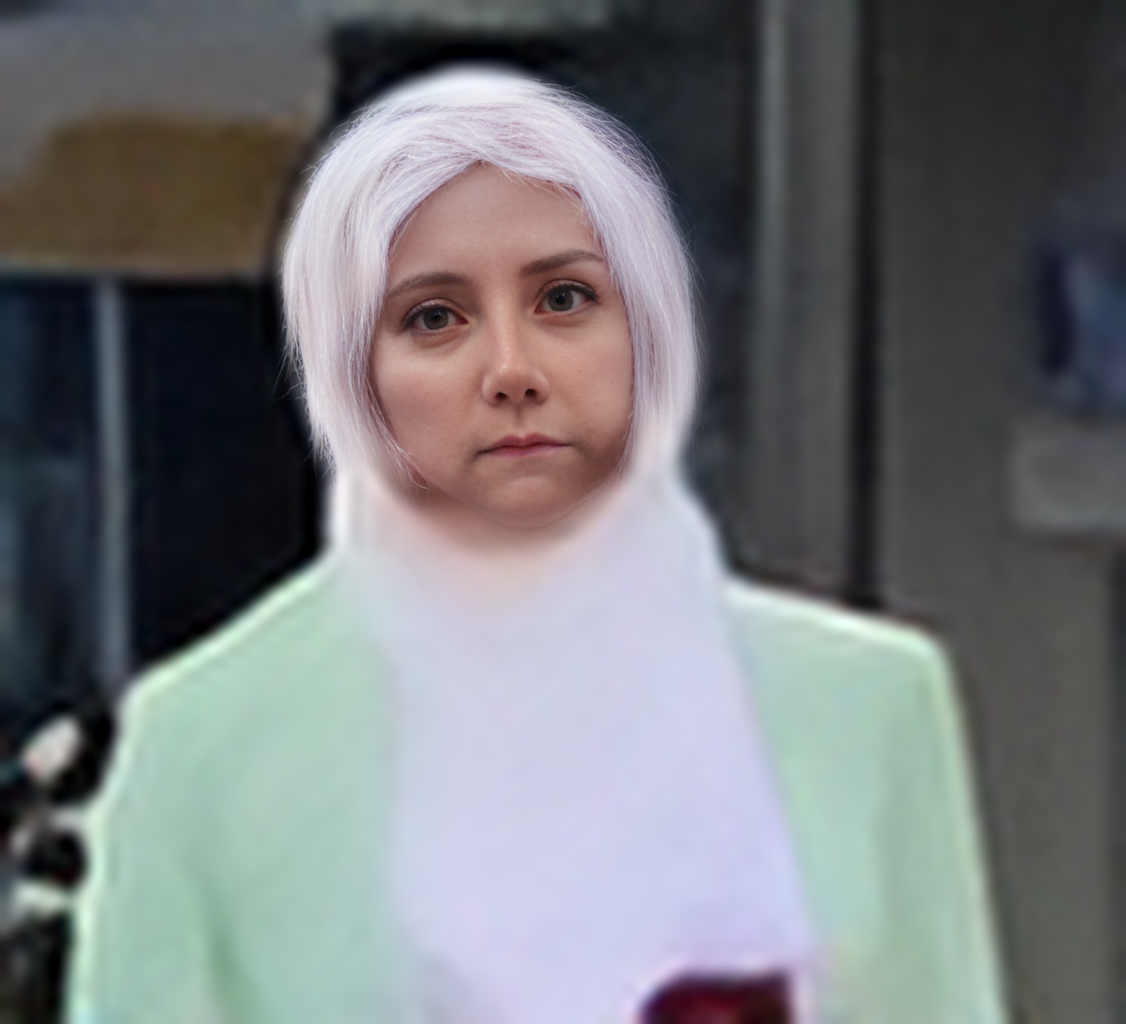} &
\includegraphics[width=\linewidth]{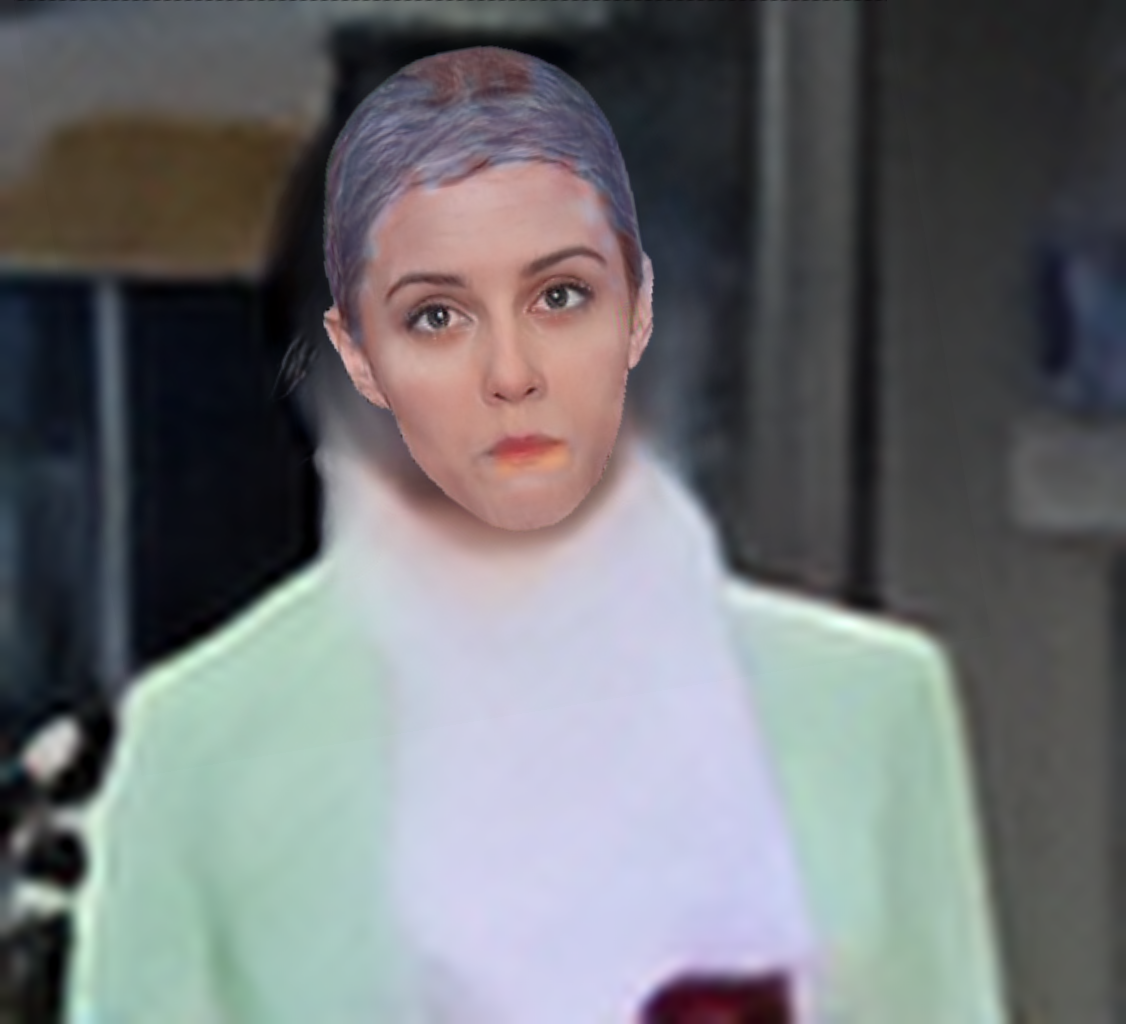} &
\includegraphics[width=\linewidth]{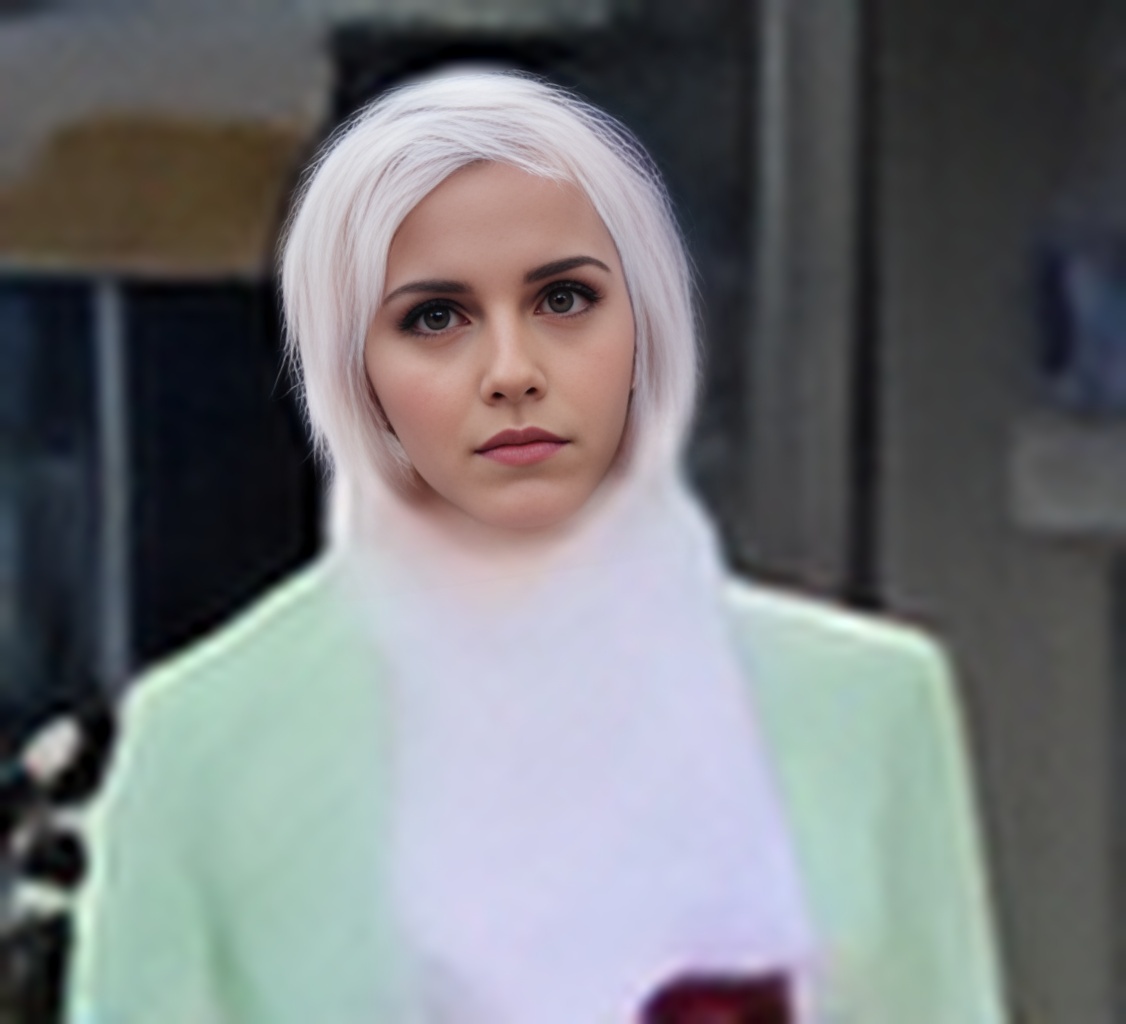} \\ \hline
\end{tabular}
\end{table}
\vspace{-5pt}\subsection{Results: Final image output from the implemented pipeline.}
\begin{figure}[H]
	\centering
	\begin{subfigure}{0.53\linewidth}
		\includegraphics[width=\linewidth]{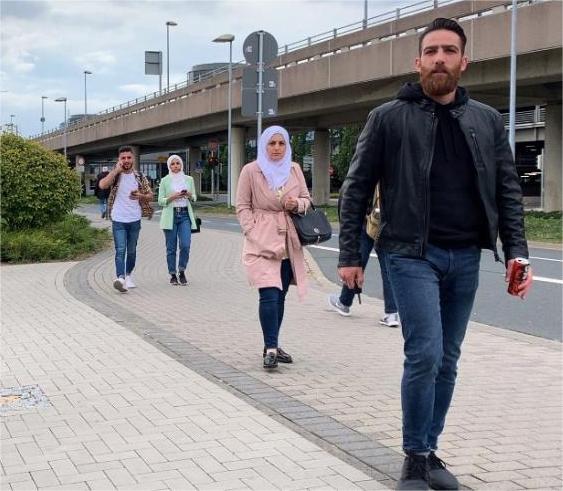}
	\end{subfigure}
    \begin{subfigure}{0.53\linewidth}
		\includegraphics[width=\linewidth]{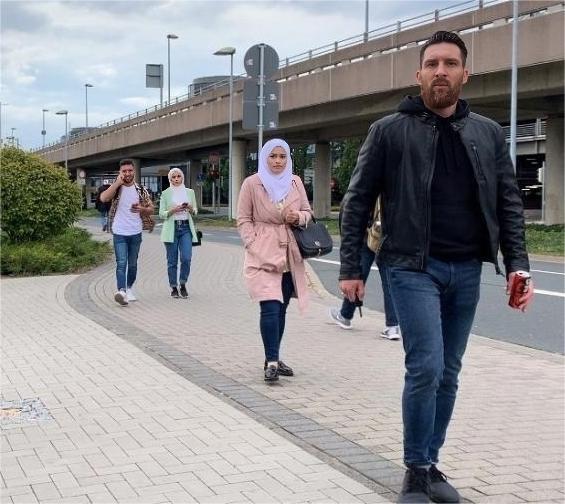}
	\end{subfigure}
    \caption{Sample output after applying the five-stage pipeline.}
	\label{tab:s1}
\end{figure}

The previous result sets were meant to signify the vitality of each module of the pipeline. This set will show the results of applying the 5 stages of the pipeline to an image taken from Egyptian streets with multiple pedestrians using Roop model since it proved to be more reliable than Ghost-v2. It is worth mentioning that the process is completely automated where the input is any street image and the output is the same image will all detectable pedestrians anonymized. Figure \ref{tab:s1} is a sample having four pedestrians at different distances from the capturing lens. The pipeline was able to detect and conceal the identity of the four pedestrians in the image.
\vspace{-5pt}\subsection{Extra Phase: Usability Testing on Subsequent Task}
\begin{figure}[H]
	\centering
	\begin{subfigure}{0.53\linewidth}
		\includegraphics[width=\linewidth]{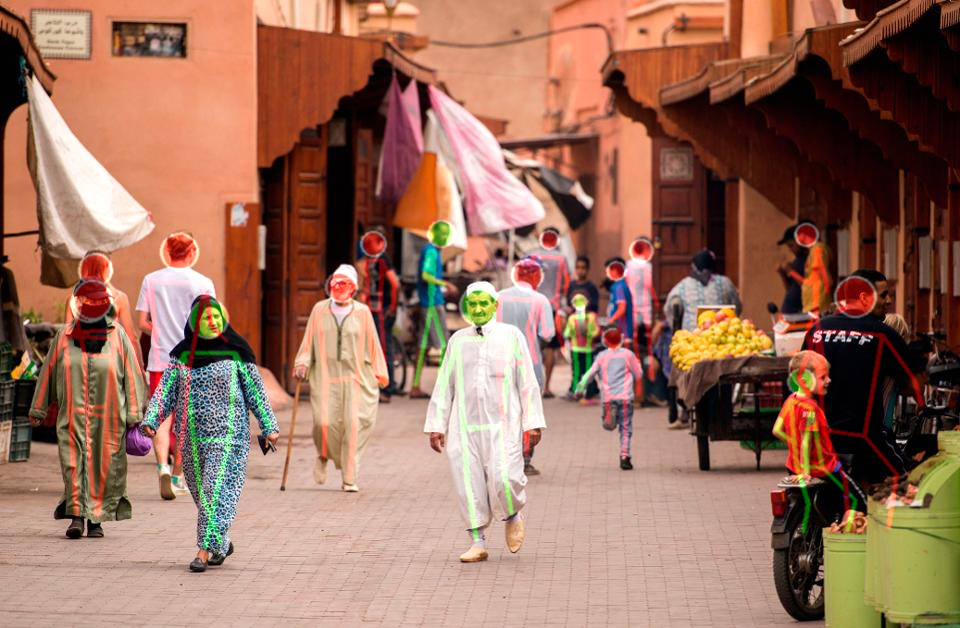}
	\end{subfigure}
    \begin{subfigure}{0.53\linewidth}
		\includegraphics[width=\linewidth]{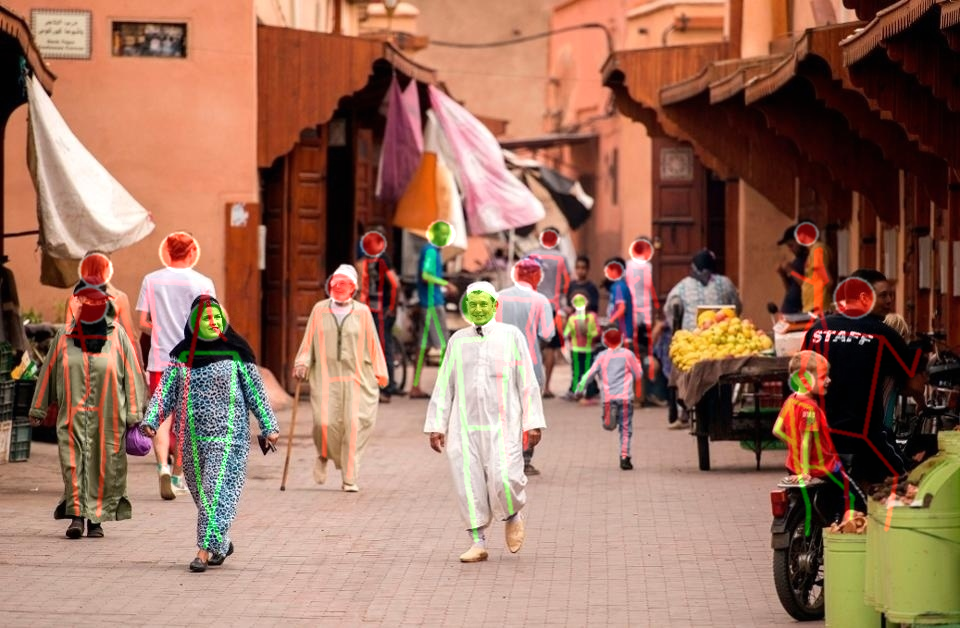}
	\end{subfigure}
    \caption{Comparison between looking/not looking feature before and after using the pipeline.}
    \label{tab:looking}
\end{figure}

Looking/ not looking feature, based on eye gaze direction, is the most crucial feature used in pedestrian intention prediction models. It is annotated in datasets as JAAD dataset \cite{JAAD}. Table \ref{tab:looking} shows the results of a pretrained feature extractor before and after applying the pipeline on all clearly visible pedestrians. This extractor highlights looking pedestrians looking to the lens with green color and the not looking ones using the red color. As shown in figure \ref{tab:looking}, all pedestrians' looking/ not looking feature is maintained from before and after the pipeline appliance which proves that the pipeline can preserve facial attributes needed in subsequent task while maintaining pedestrians privacy.
\vspace{-10pt}
\section{Conclusion and future recommendations}\label{sec6}
 Datasets involving pedestrian footage put the pedestrians’ privacy at risk. Thus, protecting pedestrians privacy is a must; it is equally important to maintain the usability of these protected images for downstream tasks. Hence, this work implemented a five-stage pipeline that aims to preserve pedestrians’ privacy while maintaining essential facial attributes especially in Egyptian streets. The results show that Roop is the perfect fit for the pipeline since it outperformed Ghost-v2 in 3 out of 4 quantitative metrics. Both models showed satisfying results on high-quality facial images, yet Roop proved to be more robust and reliable in handling challenging scenarios. Ghost-v2 failed to usually handle occluded faces and swapped the women's veil with the source hair which may be ethically inappropriate. To conclude, the implemented pipeline is successfully able to strike the required balance between pedestrians' privacy and data usability.

For future work, the pipeline is to be applied on more samples from the Egy-DRiVeS dataset to better evaluate its reliability and robustness. Additionally, It can be applied on videos if the 3-minute inference time is optimized. Moreover, the quality enhancer resizes the image to a unified size, which distorts the facial attributes when the image is resized back to its size for blending. Thus, fine-tuning the model to handle various image sizes will be beneficial. Lastly, a source face selection algorithm may elevate the performance by choosing a dissimilar source face to the target to achieve better identity concealment and to have more realistic output by accounting for age, gender, and skin tone. Also, a generative model can be used to provide completely synthetic faces for source selection.

\vspace{-10pt}
\vspace{-10pt}
\vspace{-10pt}
\vspace{-10pt}
\vspace{-10pt}
\appendices

\bibliographystyle{IEEEtran}
\bibliography{sections/ref} 

\end{document}